\pdfoutput=1

\documentclass[11pt]{article}

\usepackage[final]{acl}
\usepackage{tabularx}
\usepackage{booktabs}

\usepackage{times}
\usepackage{latexsym}
\usepackage{amsmath}

\usepackage[T1]{fontenc}

\usepackage[utf8]{inputenc}

\usepackage{microtype}

\usepackage{inconsolata}

\usepackage{graphicx}

\title{TeluguST-46: A Benchmark Corpus and Comprehensive Evaluation for Telugu-English Speech Translation
}

\author{
  \textbf{Bhavana Akkiraju\textsuperscript{1}} \quad
  \textbf{Srihari Bandarupalli\textsuperscript{1}} \quad
   \textbf{Swathi Sambangi\textsuperscript{2}} \\
  \textbf{Vasavi Ravuri\textsuperscript{2}} \quad
  \textbf{R Vijaya Saraswathi\textsuperscript{2}} \quad
  \textbf{Anil Kumar Vuppala\textsuperscript{1}} \\
  \textsuperscript{1}International Institute of Information Technology, Hyderabad, India \\
  \textsuperscript{2}VNRVJIET, Hyderabad, India \\
  \texttt{\{bhavana.akkiraju, srihari.bandarupalli\}@research.iiit.ac.in} \\
  \texttt{\{swathi\_s,vasavi\_r, vijayasaraswathi\_r\}@vnrvjiet.in} \\
  \texttt{anil.vuppala@iiit.ac.in}
}

\begin{document}
\maketitle
\begin{abstract}
Despite Telugu being spoken by over 80 million people, speech translation research for this morphologically rich language remains severely underexplored. We address this gap by developing a high-quality Telugu--English speech translation benchmark from 46 hours of manually verified CSTD corpus data (30h/8h/8h train/dev/test split). Our systematic comparison of cascaded versus end-to-end architectures shows that while IndicWhisper + IndicMT achieves the highest performance due to extensive Telugu-specific training data, fine-tuned SeamlessM4T models demonstrate remarkable competitiveness despite using significantly less Telugu-specific training data. This finding suggests that with careful hyperparameter tuning and sufficient parallel data (potentially less than 100 hours), end-to-end systems can achieve performance comparable to cascaded approaches in low-resource settings. Our metric reliability study evaluating BLEU, METEOR, ChrF++, ROUGE-L, TER, and BERTScore against human judgments reveals that traditional metrics provide better quality discrimination than BERTScore for Telugu-English translation. The work delivers three key contributions: a reproducible Telugu--English benchmark, empirical evidence of competitive end-to-end performance potential in low-resource scenarios, and practical guidance for automatic evaluation in morphologically complex language pairs.
\end{abstract}

\section{Introduction}

Speech Translation (ST) systems aim to convert spoken input in one language into textual output in another, combining the challenges of automatic speech recognition (ASR) and machine translation (MT). While ST has made significant progress for high-resource language pairs like English--French, low-resource languages from the Indian subcontinent, such as Telugu, remain vastly underexplored.

Telugu is spoken by over 80 million people and is the official language of the Indian states Andhra Pradesh and Telangana. Despite this large speaker base, Telugu-English ST research is hindered by the lack of high-quality speech-text parallel data and robust evaluation protocols. Telugu's agglutinative morphology and rich linguistic structure further complicate the task, often rendering standard evaluation metrics insufficient for capturing translation quality.

Recent efforts such as India's \textit{Bhaashini} initiative and the AI4Bharat consortium have significantly advanced NLP for Indian languages. Models like IndicWhisper~\cite{kumar2022indicwhisper}, IndicWav2Vec~\cite{javed2021building}, and IndicTrans~\cite{gala2023indictrans} have enabled ASR and MT for many Indic languages, including Telugu. These models benefit from extensive Telugu-specific training data, with IndicWhisper trained on over 800 hours of Telugu speech data and IndicMT having substantial exposure to Telugu-English parallel corpora. This language-specific training advantage becomes particularly important in low-resource scenarios.

Contemporary ST systems fall into two broad paradigms. Cascaded systems, which perform ASR followed by MT, benefit from modular design and can leverage language-specific models trained on larger datasets for each component. End-to-end systems, such as SeamlessM4T~\cite{seamless2023} and SONAR~\cite{Duquenne2023}, directly translate speech to text using single, integrated neural architectures. While end-to-end approaches often show advantages for high-resource language pairs due to joint optimization, their effectiveness on low-resource Indic language pairs may be limited by insufficient language-specific training data. For instance, SeamlessM4T's Telugu-English training data comprises only 466 hours compared to the 800+ hours available to IndicWhisper for Telugu ASR alone.

Equally important is the question of evaluation. While BLEU~\cite{papineni-etal-2002-bleu} remains a widely used metric, it is known to underperform on morphologically rich languages. Alternatives such as ChrF++~\cite{popovic-2015-chrf}, METEOR~\cite{banerjee-lavie-2005-meteor}, ROUGE~\cite{lin-2004-rouge}, and BERTScore~\cite{zhang2020bertscoreevaluatingtextgeneration} offer different perspectives on translation quality. TER~\cite{snover-etal-2006-study} provides another angle on edit distance. However, their reliability and discriminative power in Telugu--English ST scenarios has not been systematically studied, particularly regarding which metrics can effectively distinguish between different model performances.
\vspace{1mm}
\noindent\textbf{This paper addresses these gaps through three major contributions:} (1) We release 46 hours of manually verified Telugu--English parallel data\footnote{\href{https://drive.google.com/file/d/1Vlc0SO0MrYaPv7ZfZ_vcEWW-vs-3xniL/view?usp=sharing}{TeluguST-46}} curated from the CSTD corpus~\cite{mirishkar2021cstd}, facilitating reproducible benchmarks for ST in a low-resource Indian languages
 (2) We evaluate the performance of both cascaded (IndicWhisper+IndicMT, IndicWav2Vec+IndicMT) and end-to-end models (SeamlessM4T-v2 large, SONAR), providing empirical evidence that while cascaded approaches with extensive language-specific training data achieve superior performance, carefully fine-tuned end-to-end systems show promising competitiveness with limited data. (3) We analyze the reliability and discriminative power of six standard automatic metrics (BLEU, METEOR, ChrF++, ROUGE, TER, and BERTScore) by comparing them with ChatGPT-style human evaluations.

We also investigate the effect of mixing verified and unverified data during fine-tuning, offering insights into corpus construction strategies under resource constraints. Together, these efforts provide a solid foundation for Telugu--English ST benchmarking and offer broader implications for other morphologically rich, underrepresented languages. All resources—data splits, evaluation scripts, and system outputs—are made publicly available to foster reproducible and inclusive research in multilingual speech translation.
\begin{table*}[!htbp]
\centering
\begin{tabular}{lp{13cm}}
\toprule
\textbf{Metric} & \textbf{Description} \\
\midrule
BLEU & Focuses on lexical matching and word order. \\
ChrF++ & Captures sub-word similarities and handles morphological variations. \\
BERTScore & Captures semantic meaning beyond surface form matching. \\
TER & Measures minimum number of edits (insertions, deletions, substitutions, shifts) required to transform hypothesis into reference. \\
ROUGE-L & Captures structural similarity and word order. \\
METEOR & Incorporates linguistic knowledge and semantic equivalence. \\
\bottomrule
\end{tabular}
\caption{Overview of evaluation metrics and their calculation approaches for assessing translation quality.}
\label{tab:metric-overview}
\end{table*}
\section{Dataset Creation} We construct a high-quality Telugu-English speech translation dataset from the CSTD corpus \cite{mirishkar2021cstd}, addressing cross-lingual alignment and dialectal variation challenges through systematic quality control.
\subsection{Source Dataset Selection}
We utilize the CSTD corpus, comprising 2,000 hours of transcribed Telugu audio across three dialectal variants: Telangana, Rayalaseema, and Andhra. We strategically selected a 50-hour subset maintaining dialectal diversity while ensuring computational feasibility for manual verification processes.

\subsection{ Translation Pipeline and Quality Control}

\subsubsection{ Initial Translation Generation}
Telugu transcripts were extracted from the 50-hour subset and translated to English using IndicTrans MT, selected for its effectiveness with morphologically rich Indian languages and Telugu-English parallel training data.

\subsubsection{Manual Verification Protocol}
Each audio-translation pair underwent evaluation by native Telugu speakers using a 5-point Likert scale:
\begin{itemize}
    \item Score 5: Perfect semantic alignment with natural English expression
    \item Score 4: Minor variations, core meaning preserved
    \item Score 3: Acceptable translation with moderate modifications required
    \item Score 2: Significant gaps requiring substantial 
    revision
    \item Score 1: Fundamental misalignment requiring complete retranslation

\end{itemize}

Pairs scoring below 3 received manual correction. Audio segments with multiple speakers, background noise, or recording artifacts were eliminated through automatic detection and manual inspection.

\subsubsection{Dataset Composition}
Following quality control, we obtained 46 hours of verified parallel data (92\% retention rate), partitioned as Training Set: 30 hours (65.2\%),Development Set: 8 hours (17.4\%), Test Set: 8 hours (17.4\%)

\subsubsection{ Key Challenges}
Primary challenges included transcription inconsistencies from dialectal variations, translation ambiguities from Telugu's morphological richness, and temporal alignment drift. These were addressed through standardization protocols, manual correction procedures, and systematic alignment verification.

\section{Speech Translation Architectures and Training Setup}
\subsection{Model Architectures}

We evaluate four speech translation architectures spanning two major paradigms: cascaded and end-to-end systems.

The cascaded models follow a modular design and benefit from component-specific training advantages. \textbf{IndicWhisper + IndicMT} (cascad-1) integrates IndicWhisper, an ASR model fine-tuned on over 800 hours of Telugu speech data, with IndicTrans, a neural machine translation system with extensive Telugu-English parallel training. This combination leverages substantial language-specific training data for both ASR and MT components. \textbf{IndicWav2Vec + IndicMT} (cascad-2) combines a Telugu-specific Wav2Vec-based speech encoder with IndicTrans. However, this configuration faces limitations as the Wav2Vec model~\footnote{https://huggingface.co/anuragshas/wav2vec2-large-xlsr-53-telugu} has been trained on significantly limited Telugu data compared to IndicWhisper, potentially leading to incorrect transcriptions that negatively impact downstream translation quality.

The end-to-end models include \textbf{SeamlessM4T} and \textbf{SONAR}. SeamlessM4T is a massively multilingual model that directly maps speech to text using a single, integrated architecture, but with limited Telugu-English training data (466 hours) compared to the cascaded components. Despite this limitation, our experiments reveal that SeamlessM4T shows promising potential when fine-tuned with careful hyperparameter selection. SONAR is a unified multilingual encoder that produces sentence-level representations for over 200 languages across both speech and text modalities, though it similarly lacks the extensive Telugu-specific training of the cascaded components.

\subsection{Training Configurations}

We investigate the impact of \textbf{training data quality and quantity} by fine-tuning \textit{SeamlessM4T-v2-large} under several data configurations. Three primary data setups are defined:

\begin{itemize}
    \item \textbf{Verified Only:} 30 hours of manually verified, high-quality speech–text pairs used as the clean-data baseline.
    \item \textbf{Balanced:} 45 hours of data combining 30 hours of verified samples with an additional 15 hours of unverified speech–text pairs.
    \item \textbf{Unbalanced:} 73–80 hours of data with a larger proportion of unverified samples, allowing us to examine the effects of noisy data on model robustness.
\end{itemize}

For \textbf{hyperparameter optimization}, we explore learning rates of $1 \times 10^{-5}$ and $1 \times 10^{-6}$, batch sizes of 10, 32, and 64, and warmup steps of 100. These settings are used to identify stable training regimes that balance learning efficiency and generalization performance across the various data configurations.

We define six fine-tuning configurations as follows:

\begin{itemize}
    \item \textbf{Configuration 0:} Direct inference using the pretrained \textit{SeamlessM4T-v2-large} model without fine-tuning, serving as the baseline.
    
    \item \textbf{Configuration 1 (Verified Only):} Fine-tuning on 30 hours of verified data to assess model performance under clean training conditions.
    
    \item \textbf{Configuration 2 (Balanced):} Fine-tuning on 45 hours of mixed data (30 verified + 15 unverified) to evaluate the benefits of moderate data augmentation.
    
    \item \textbf{Configurations 3--5: Extended Mixed Data (80 hours)} \\
    Employed a larger dataset combining verified and unverified data (80 hours total) to investigate the effects of hyperparameter variations at scale.    
   
    \textbf{Configuration 3:} Learning rate = $1 \times 10^{-5}$, batch size = 10, representing an aggressive learning setup with smaller batch processing.
   
    \textbf{Configuration 4:} Learning rate = $1 \times 10^{-6}$, batch size = 10, combining conservative learning with smaller batch sizes to balance stability and computational efficiency.   
    
    \textbf{Configuration 5:} Learning rate = $1 \times 10^{-6}$, batch size = 32, implementing a conservative learning strategy with larger batch size for improved gradient stability.
\end{itemize}

This setup allows us to comprehensively assess the relationship between data quality, quantity, and hyperparameter sensitivity in fine-tuning SeamlessM4T for Telugu–English speech translation.

\begin{table*}[!htbp]
\setlength{\tabcolsep}{1.5pt} 
\centering
\begin{tabular}{lccccccc}
\hline
\textbf{Model} & \textbf{BLEU} & \textbf{ChrF++} & \textbf{TER (\textdownarrow)} & \textbf{ROUGE-L} & \textbf{METEOR} & \textbf{BERTScore} & \textbf{ChatGPT (\%)}\\
\hline
\textbf{cascad-1} & \textbf{24.0} & \textbf{47.7} & \textbf{82.0} & \textbf{48.5} & \textbf{47.2} & 91.9 & \textbf{58.3}\\
cascad-2 & 8.3 & 28.3 & 99.8 & 28.2 & 20.2 & 89.1 & 41.2\\
\hline
Configuration 0 & 13.9 & 38.9 & 89.9 & 38.7 & 39.5 & 89.5 & 48.1\\
configuration 1  & 14.1 & 39.8 & 89.5 & 40.2 & 40.2 & 89.1 & 49.2 \\
configuration 2  & 16.9 & 42.87 & 86.0 & 42.6 & 43.01 & 90.1 & 51.0 \\
configuration 3& 14.9 & 40.1 & 87.7 & 41.2 & 41.6 & 89.7 & 50.4\\
configuration 4 & 17.1 & 42.9 & 85.9 & 42.7 & 43.8 & 90.9  & 51.3\\
configuration 5 & 15.8 & 41.8 & 86.0 & 41.6 & 42.4 & 89.9 & 50.9\\
SONAR & 10.9 & 35.8 & 98.7 & 36.4 & 35.6 & 88.3 &49.2 \\
\hline
\end{tabular}
\caption{Automatic metric scores across cascaded and end-to-end models. Cascaded models, particularly IndicWhisper + IndicMT, significantly outperform end-to-end approaches across most metrics.}
\label{tab:results-automatic}
\end{table*}

\section{Evaluation Metrics}

We employ six automatic metrics capturing different translation quality aspects. Table~\ref{tab:metric-overview} provides an overview of these metrics and their computational approaches, highlighting how each metric addresses different dimensions of translation quality assessment.

\subsection{BLEU Score}
Measures precision-based n-gram overlap with brevity penalty, emphasizing lexical similarity.
\begin{equation}
\text{BLEU} = BP \cdot \exp\left(\sum_{n=1}^{N} w_n \log p_n\right)
\end{equation}
where $p_n$ is n-gram precision and $BP$ is brevity penalty.

\subsection{ChrF++}
Evaluates character-level F-score, effective for morphologically rich languages like Telugu.
\begin{equation}
\text{ChrF++} = \frac{(1+\beta^2) \cdot chrP \cdot chrR}{\beta^2 \cdot chrP + chrR}
\end{equation}
where $chrP$ and $chrR$ are character-level precision and recall.

\subsection{BERTScore}
Computes semantic similarity using contextual embeddings, capturing meaning beyond surface form.
\begin{equation}
\text{BERTScore} = \frac{1}{|x|} \sum_{x_i \in x} \max_{y_j \in y} \mathbf{x}_i^\top \mathbf{y}_j
\end{equation}
where $\mathbf{x_i}$ and $\mathbf{y_j}$ are BERT embeddings of tokens.

\subsection{Translation Edit Rate (TER)}
Measures minimum edit operations required for transformation (lower is better).
\begin{equation}
\text{TER} = \frac{\text{Number of Edits}}{\text{Average Reference Length}}
\end{equation}

\subsection{ROUGE}
Computes recall-oriented n-gram overlap. We use ROUGE-1, ROUGE-2, and ROUGE-L variants.
\begin{equation}
\text{ROUGE-N} = \frac{\text{Number of matching n-grams}}{\text{Total n-grams in the reference}}
\end{equation}
\begin{equation}
\text{ROUGE-L} = \frac{(1+\beta^2) \cdot P \cdot R}{\beta^2 \cdot P + R}
\end{equation}
where $P$ is precision, $R$ is recall, and $\beta = 1$.

\subsection{METEOR}
Combines precision and recall using exact, stem, synonym, and paraphrase matching with linguistic equivalence.
\begin{equation}
\text{METEOR} = \frac{10 \cdot P \cdot R}{R + 9 \cdot P} \cdot (1 - \text{Penalty})
\end{equation}
where $P$ and $R$ are weighted precision and recall, and $Penalty$ accounts for word order differences.


\section{Experimental Setup}
\subsection{Baseline Model Evaluation}
Our experimental framework comprised two distinct categories of speech translation models: cascaded and end-to-end architectures. For the cascaded approaches, we evaluated two configurations: IndicWhisper + IndicMT and IndicWav2Vec + IndicMT. Additionally, we assessed \textbf{SONAR} and \textbf{Seamless} as our end-to-end baseline models. Initial performance evaluation was conducted by directly inferencing these models on our designated test dataset without any domain-specific adaptations. This provided baseline performance metrics that informed subsequent fine-tuning strategies.

\subsection{Fine-Tuning Methodology}
Based on the initial inference results (Configuration 0 in Table~\ref{tab:results-automatic}, which showed end-to-end models underperforming cascaded approaches, we implemented a systematic fine-tuning approach~\cite{akkiraju-etal-2025-iiith} using the SeamlessM4T model to investigate whether targeted fine-tuning could bridge this performance gap . The fine-tuning process was designed to investigate the impact of training data composition and hyperparameter configurations on model performance.

\section{Results}

We evaluate cascaded and end-to-end speech translation models using six automatic metrics—BLEU, ChrF++, TER, ROUGE-L, METEOR, and BERTScore—alongside ChatGPT-style human scoring to assess model performance and metric reliability for Telugu--English speech translation. Table~\ref{tab:results-automatic} presents results across all configurations.

Our findings highlight the performance dynamics across architectural choices. \textbf{IndicWhisper + IndicMT} achieves the best overall performance with 23.99 BLEU, 47.69 ChrF++, and 47.19 METEOR, clearly benefiting from over 800 hours of Telugu ASR training and substantial Telugu-English parallel corpora in the MT stage. This cascaded setup demonstrates the importance of component-specific language exposure in low-resource scenarios.

Fine-tuned SeamlessM4T models, however, demonstrate competitive performance despite limited Telugu-specific supervision (see Table~\ref{tab:results-automatic}).\textbf{Configuration 4} achieves 17.1 BLEU and 43.82 METEOR—only a 29\% BLEU drop from the cascaded baseline—while using under 30 hours of verified training data and 43 hours of unverified training data. This result suggests that with effective fine-tuning, end-to-end models can approach cascaded system performance, offering a scalable alternative where monolithic training pipelines are preferred.

By contrast, \textbf{Wav2Vec + IndicMT} underperforms across all metrics (8.3 BLEU, 28.27 ChrF++, 20.17 METEOR), largely due to inadequate Telugu ASR training data. This discrepancy underscores the vital role of ASR quality in cascaded speech translation pipelines. Additionally, \textbf{SONAR} achieves only moderate performance (10.9 BLEU, 35.6 METEOR), highlighting that not all end-to-end multilingual systems generalize well to morphologically rich Indian languages. Among end-to-end SeamlessM4T variants, \textbf{Configurations 2 and 4} deliver the best results. Configuration 2 achieves 16.9 BLEU and 43.01 METEOR, closely trailing Configuration 4. These results reflect the effectiveness of balanced data curation and learning rate tuning in maximizing translation quality with limited resources.

\begin{table}[ht]
\centering
\begin{tabular}{lc}
\toprule
\textbf{Metric} & \textbf{PCC with ChatGPT} \\
\midrule
BLEU       & 0.94 \\
ChrF++     & 0.96 \\
TER        & -0.82 \\
ROUGE-L    & 0.97 \\
METEOR     & 0.90 \\
BERTScore  & 0.73 \\
\bottomrule
\end{tabular}
\caption{Pearson correlation (PCC) between ChatGPT-style scores and automatic metrics across model configurations.}
\label{tab:chatgpt-metric-corr}
\end{table}

\subsection{Metric Reliability Analysis}

To evaluate which metrics align best with human perception, we compute the Pearson correlation coefficient (PCC) between ChatGPT scores and each automatic metric.

\begin{itemize}
    \item \textbf{ROUGE-L ($r = 0.97$)} and \textbf{ChrF++ ($r = 0.96$)} show the strongest agreement with human judgment, making them suitable for low-resource ST evaluation.
    \item \textbf{BLEU ($r = 0.94$)} and \textbf{METEOR ($r = 0.90$)} also show strong correlations, reinforcing their applicability in this domain.
    \item \textbf{BERTScore ($r = 0.73$)} shows lower alignment across configurations, suggesting limited discriminative power for morphologically rich translations.
    \item \textbf{TER ($r = -0.82$)} exhibits a strong inverse correlation, as expected from an error-based metric, further validating its interpretability in this context.
\end{itemize}

\section{Conclusion}

This work presents a comprehensive benchmark for Telugu--English speech translation, releasing 46 hours of verified data from the CSTD corpus. Our experiments reveal important insights into the trade-offs between cascaded and end-to-end architectures in low-resource settings. While cascaded systems like IndicWhisper + IndicMT achieve the best results (23.99 BLEU, 47.19 METEOR), fine-tuned end-to-end models such as SeamlessM4T demonstrate notable competitiveness (17.1 BLEU, 43.82 METEOR) despite relying on significantly less Telugu-specific training data.

This 29\% BLEU gap underscores the strong inductive bias provided by Telugu-specific training in cascaded models. However, it also highlights the scalability potential of end-to-end systems: with less than 100 hours of parallel fine-tuning data, SeamlessM4T models can approach the performance of cascaded systems. The weak performance of Wav2Vec + IndicMT (8.3 BLEU, 28.27 ChrF++) further emphasizes the necessity of sufficient ASR exposure for effective cascaded ST.

Our metric reliability analysis demonstrates that \textbf{ROUGE-L} and \textbf{ChrF++} correlate most strongly with human judgments (Pearson $r = 0.97$ and $r = 0.96$, respectively), suggesting their effectiveness in evaluating low-resource, morphologically rich translations. \textbf{BLEU} and \textbf{METEOR} also show strong alignment ($r > 0.90$), reinforcing their continued relevance in this setting. In contrast, \textbf{BERTScore} exhibits lower correlation ($r = 0.73$), indicating limited sensitivity to performance variation across configurations. As expected, \textbf{TER} shows strong inverse correlation ($r = -0.82$), aligning well with human judgments when interpreted as an error-based metric.

Although cascaded systems benefit from large language-specific resources (e.g., 800+ hours for IndicWhisper), our findings indicate that carefully fine-tuned end-to-end systems, with modest high-quality data, can deliver competitive performance. Future work should explore monolingual augmentation, domain-specific adaptation (e.g., medical ST), and multilingual pretraining across related Indian languages to further enhance end-to-end ST in underrepresented contexts.

Overall, this benchmark lays the foundation for reproducible research in Indic speech translation, encourages deeper evaluation methodology, and demonstrates that with thoughtful design, end-to-end systems offer a viable path forward in low-resource speech translation scenarios.

\section{Limitations and Future Work}

\subsection{Limitations}

Our study presents few  limitations that should be considered when interpreting the results. The primary limitation is the constrained dataset size of 46 hours, which, while manually verified for quality, limits our ability to fully understand the relationship between data quantity and model performance. This constraint prevents us from conducting comprehensive scaling experiments to determine optimal data requirements for different model architectures. The potential for including more verified data from the CSTD corpus remains unexplored, which could provide valuable insights into how data size impacts translation quality across cascaded and end-to-end approaches. Another significant limitation is our focus on a single end-to-end model (SeamlessM4T) for fine-tuning experiments. While this provided insights into the potential of end-to-end approaches, exploring other architectures could yield different conclusions about the scalability and effectiveness of end-to-end speech translation systems for Telugu.

\subsection{Future Work}

Several promising research directions emerge from this work that could significantly advance Telugu speech translation and broader low-resource ST research.

\textbf{Dataset Expansion and Quality Analysis:} A critical next step involves scaling the verified dataset to investigate the data size-performance relationship systematically. Expanding from the current 46 hours to 100+ hours of verified Telugu-English parallel data would enable more robust conclusions about optimal training data requirements for different model architectures.

\textbf{Advanced Fine-tuning Strategies for End-to-End Models:} Given the promising results of SeamlessM4T fine-tuning, future research should explore more sophisticated adaptation techniques including domain adaptation strategies, few-shot learning approaches, and parameter-efficient fine-tuning methods such as LoRA or adapters.

\textbf{Cross-lingual Transfer and Multilingual Extensions:} Investigating how insights from Telugu-English speech translation can transfer to other Dravidian languages (Tamil, Kannada, Malayalam) could provide valuable guidance for developing ST systems across the Indian subcontinent. This includes exploring multilingual training strategies and cross-lingual transfer learning approaches that leverage shared linguistic properties among related languages.

\bibliography{custom}

@inproceedings{mirishkar2021cstd,
  title={CSTD-Telugu corpus: Crowd-sourced approach for large-scale speech data collection},
  author={Mirishkar, Ganesh S and Naroju, Meher Dinesh and Maity, Sudhamay and Yalla, Prakash and Vuppala, Anil Kumar and others},
  booktitle={2021 Asia-Pacific Signal and Information Processing Association Annual Summit and Conference (APSIPA ASC)},
  pages={511--517},
  year={2021},
  organization={IEEE}
}

@inproceedings{seamless2023,
   title="Seamless: Multilingual Expressive and Streaming Speech Translation",
   author="{{Seamless Communication}} and Lo{\"i}c Barrault and Yu-An Chung and Mariano Coria Meglioli and David Dale and Ning Dong and Mark Duppenthaler and Paul-Ambroise Duquenne and Brian Ellis and Hady Elsahar and Justin Haaheim and John Hoffman and Min-Jae Hwang and Hirofumi Inaguma and Christopher Klaiber and Ilia Kulikov and Pengwei Li and Daniel Licht and Jean Maillard and Ruslan Mavlyutov and Alice Rakotoarison and Kaushik Ram Sadagopan and Abinesh Ramakrishnan and Tuan Tran and Guillaume Wenzek and Yilin Yang and Ethan Ye and Ivan Evtimov and Pierre Fernandez and Cynthia Gao and Prangthip Hansanti and Elahe Kalbassi and Amanda Kallet and Artyom Kozhevnikov and Gabriel Mejia and Robin San Roman and Christophe Touret and Corinne Wong and Carleigh Wood and Bokai Yu and Pierre Andrews and Can Balioglu and Peng-Jen Chen and Marta R. Costa-juss{\`a} and Maha Elbayad and Hongyu Gong and Francisco Guzm{\'a}n and Kevin Heffernan and Somya Jain and Justine Kao and Ann Lee and Xutai Ma and Alex Mourachko and Benjamin Peloquin and Juan Pino and Sravya Popuri and Christophe Ropers and Safiyyah Saleem and Holger Schwenk and Anna Sun and Paden Tomasello and Changhan Wang and Jeff Wang and Skyler Wang and Mary Williamson",
  journal={ArXiv},
  year={2023}
}

@misc{kumar2022indicwhisper,
      title={Vistaar: Diverse Benchmarks and Training Sets for Indian Language ASR}, 
      author={Kaushal Santosh Bhogale and Sai Sundaresan and Abhigyan Raman and Tahir Javed and Mitesh M. Khapra and Pratyush Kumar},
      year={2023},
      eprint={2305.15386},
      archivePrefix={arXiv},
      primaryClass={cs.CL},
      url={https://arxiv.org/abs/2305.15386}, 
}

@inproceedings{javed2021building,
    title = {Towards Building ASR Systems for the Next Billion Users},
    author = {Tahir Javed and Sumanth Doddapaneni and Abhigyan Raman and Kaushal Santosh Bhogale and Gowtham Ramesh and Anoop Kunchukuttan and Pratyush Kumar and Mitesh M. Khapra},
    booktitle = "Proceedings of the AAAI Conference on Artificial Intelligence",
    year = "2022",
}

@misc{Duquenne2023,
  author = {Paul-Ambroise Duquenne and Holger Schwenk and Benoit Sagot},
  title = {{SONAR:} Sentence-Level Multimodal and Language-Agnostic Representations},
  publisher = {arXiv},
  year = {2023},
  url = {https://arxiv.org/abs/2308.11466},
}

@article{gala2023indictrans,
title={IndicTrans2: Towards High-Quality and Accessible Machine Translation Models for all 22 Scheduled Indian Languages},
author={Jay Gala and Pranjal A Chitale and A K Raghavan and Varun Gumma and Sumanth Doddapaneni and Aswanth Kumar M and Janki Atul Nawale and Anupama Sujatha and Ratish Puduppully and Vivek Raghavan and Pratyush Kumar and Mitesh M Khapra and Raj Dabre and Anoop Kunchukuttan},
journal={Transactions on Machine Learning Research},
issn={2835-8856},
year={2023},
url={https://openreview.net/forum?id=vfT4YuzAYA},
note={}
}

@inproceedings{papineni-etal-2002-bleu,
    title = "{B}leu: a Method for Automatic Evaluation of Machine Translation",
    author = "Papineni, Kishore  and
      Roukos, Salim  and
      Ward, Todd  and
      Zhu, Wei-Jing",
    editor = "Isabelle, Pierre  and
      Charniak, Eugene  and
      Lin, Dekang",
    booktitle = "Proceedings of the 40th Annual Meeting of the Association for Computational Linguistics",
    month = jul,
    year = "2002",
    address = "Philadelphia, Pennsylvania, USA",
    publisher = "Association for Computational Linguistics",
    url = "https://aclanthology.org/P02-1040/",
    doi = "10.3115/1073083.1073135",
    pages = "311--318"
}

@inproceedings{popovic-2015-chrf,
    title = "chr{F}: character n-gram {F}-score for automatic {MT} evaluation",
    author = "Popovi{\'c}, Maja",
    editor = "Bojar, Ond{\v{r}}ej  and
      Chatterjee, Rajan  and
      Federmann, Christian  and
      Haddow, Barry  and
      Hokamp, Chris  and
      Huck, Matthias  and
      Logacheva, Varvara  and
      Pecina, Pavel",
    booktitle = "Proceedings of the Tenth Workshop on Statistical Machine Translation",
    month = sep,
    year = "2015",
    address = "Lisbon, Portugal",
    publisher = "Association for Computational Linguistics",
    url = "https://aclanthology.org/W15-3049/",
    doi = "10.18653/v1/W15-3049",
    pages = "392--395"
}

@misc{zhang2020bertscoreevaluatingtextgeneration,
      title={BERTScore: Evaluating Text Generation with BERT}, 
      author={Tianyi Zhang and Varsha Kishore and Felix Wu and Kilian Q. Weinberger and Yoav Artzi},
      year={2020},
      eprint={1904.09675},
      archivePrefix={arXiv},
      primaryClass={cs.CL},
      url={https://arxiv.org/abs/1904.09675}, 
}

@inproceedings{banerjee-lavie-2005-meteor,
    title = "{METEOR}: An Automatic Metric for {MT} Evaluation with Improved Correlation with Human Judgments",
    author = "Banerjee, Satanjeev  and
      Lavie, Alon",
    editor = "Goldstein, Jade  and
      Lavie, Alon  and
      Lin, Chin-Yew  and
      Voss, Clare",
    booktitle = "Proceedings of the {ACL} Workshop on Intrinsic and Extrinsic Evaluation Measures for Machine Translation and/or Summarization",
    month = jun,
    year = "2005",
    address = "Ann Arbor, Michigan",
    publisher = "Association for Computational Linguistics",
    url = "https://aclanthology.org/W05-0909/",
    pages = "65--72"
}

@inproceedings{lin-2004-rouge,
    title = "{ROUGE}: A Package for Automatic Evaluation of Summaries",
    author = "Lin, Chin-Yew",
    booktitle = "Text Summarization Branches Out",
    month = jul,
    year = "2004",
    address = "Barcelona, Spain",
    publisher = "Association for Computational Linguistics",
    url = "https://aclanthology.org/W04-1013/",
    pages = "74--81"
}

@inproceedings{snover-etal-2006-study,
    title = "A Study of Translation Edit Rate with Targeted Human Annotation",
    author = "Snover, Matthew  and
      Dorr, Bonnie  and
      Schwartz, Rich  and
      Micciulla, Linnea  and
      Makhoul, John",
    booktitle = "Proceedings of the 7th Conference of the Association for Machine Translation in the Americas: Technical Papers",
    month = aug # " 8-12",
    year = "2006",
    address = "Cambridge, Massachusetts, USA",
    publisher = "Association for Machine Translation in the Americas",
    url = "https://aclanthology.org/2006.amta-papers.25/",
    pages = "223--231",
}

@inproceedings{akkiraju-etal-2025-iiith,
    title = "{IIITH}-{BUT} system for {IWSLT} 2025 low-resource {B}hojpuri to {H}indi speech translation",
    author = "Akkiraju, Bhavana  and
      Pothula, Aishwarya  and
      Kesiraju, Santosh  and
      Vuppala, Anil",
    editor = "Salesky, Elizabeth  and
      Federico, Marcello  and
      Anastasopoulos, Antonis",
    booktitle = "Proceedings of the 22nd International Conference on Spoken Language Translation (IWSLT 2025)",
    month = jul,
    year = "2025",
    address = "Vienna, Austria (in-person and online)",
    publisher = "Association for Computational Linguistics",
    url = "https://aclanthology.org/2025.iwslt-1.34/",
    pages = "333--339",
    ISBN = "979-8-89176-272-5"
}

\end{document}